# Setting an attention region for convolutional neural networks using region selective features, for recognition of materials within glass vessels

Sagi Eppel[1]

## Abstract

Convolutional neural networks have emerged as the leading method for the classification and segmentation of images. In some cases, it is desirable to focus the attention of the net on a specific region in the image; one such case is the recognition of the contents of transparent vessels, where the vessel region in the image is already known. This work presents a valve filter approach for focusing the attention of the net on a region of interest (ROI). In this approach, the ROI is inserted into the net as a binary map. The net uses a different set of convolution filters for the ROI and background image regions, resulting in a different set of features being extracted from each region. More accurately, for each filter used on the image, a corresponding valve filter exists that acts on the ROI map and determines the regions in which the corresponding image filter will be used. This valve filter effectively acts as a valve that inhibits specific features in different image regions according to the ROI map. In addition, a new data set for images of materials in glassware vessels in a chemistry laboratory setting is presented. This data set contains a thousand images with pixel-wise annotation according to categories ranging from filled and empty to the exact phase of the material inside the vessel. The results of the valve filter approach and fully convolutional neural nets (FCN) with no ROI input are compared based on this data set.

## 1. Introduction

Deep learning methods based on convolutional neural networks (CNN) have transformed the field of computer vision, allowing computers to achieve near-human level results in tasks such as image classification, object detection and semantic segmentation [1-4]. In many cases, it is desirable to restrict the recognition problem to a given region of the image. This region is often referred to as an attention mask or a region of interest (ROI). In many cases, the ROI is known and is given as an input to the net (Figure 1a). In such cases, the challenge is to focus the attention of the net on the ROI without loss of background information. One such case is the identification of the contents of transparent vessels such as bottles, jars and other glassware, in which the glassware region in the image is known and is given as an ROI input (Figure 1a). This task is essential in a variety of fields, ranging from chemistry laboratory work to everyday beverage handling [5-9]. The general problem of the recognition and segmentation of bottles, jars and other glassware is already covered by a variety of existing nets and data sets (COCO, KADE20) [10-12]. However, the more specific task of recognition of the vessel's contents is not covered by any existing data set or currently available neural nets.

---

[1] Vayavision, sagieppel@gmail.com

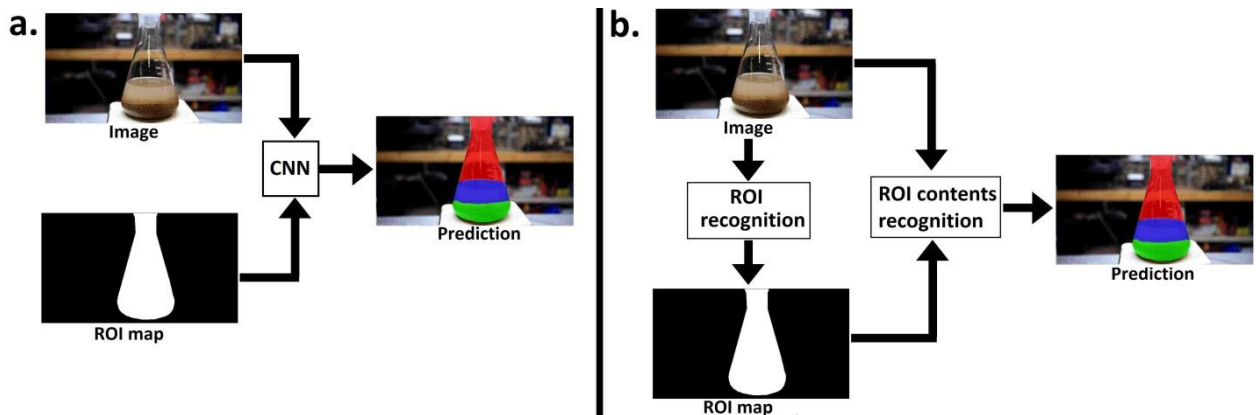

**Figure 1. a) Convolutional neural net (CNN) with ROI map as input; b) Hierarchical segmentation. The glassware region is found by a first method and is used as ROI input for the vessel content recognition net.**

It is therefore appropriate to use an existing method for the general problem of finding the region of the glassware in the image (Figure 1b). The output of this net can then be used as ROI input for another net, which specializes in the specific task of recognizing the vessel's contents (Figure 1b). The alternative option of using a single net to identify the vessel and its contents (Figure 2) is also examined in this work; however, this gives results which are inferior to ROI-based methods.

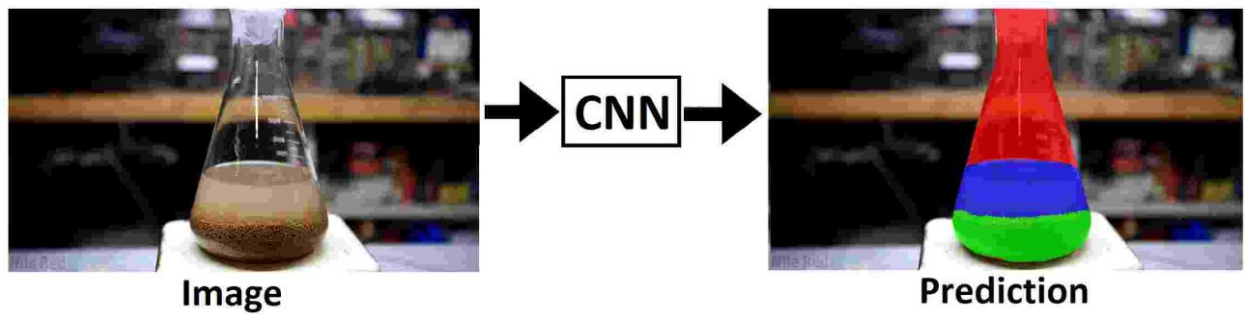

**Figure 2. Recognition of the vessel region and its contents using a single net**

When using a convolutional neural net on an image for which a region of interest (ROI) map is given as an input (Figure 1a), it makes sense to search for different kinds of features in different regions [13]. Convolutional neural nets extract features from the image by convolving a set of different filters with the image to create a feature map for each filter [3]; hence, it is desirable to apply a different set of filters to the background and ROI regions. The approach suggested here is based on applying the same set of filters to the entire image. However, some features are later inhibited in image regions corresponding to the ROI or background, using the valve filter approach. The valve filters are applied to the ROI map (Figure 3) using a simple convolution to give a relevance map, in the same manner that the image filters are applied to the image to give a feature map (Figure 3). The feature map created by the image filter is multiplied (element-wise) by the relevance map created by the corresponding valve filter, to give a normalized feature map that is passed as input to the next layer of the net (Figure 3). In this way, the valve filters acts as a kind of valve that inhibits specific filters in the ROI or background regions. Hence, all filters are applied to the entire image and the valve filters inhibit different features in different image regions and cause an effect which is similar to extracting different features from different regions.

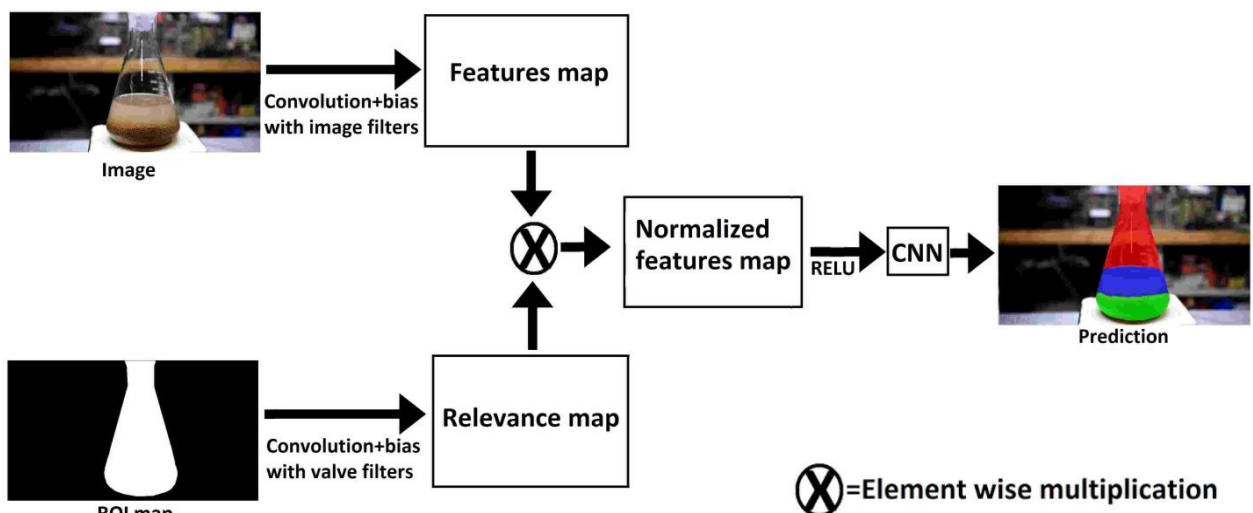

**Figure 3.** The valve filter approach for introduction of the ROI map as input to convolutional neural nets. The image and the ROI input are each passed through a separate convolution layer to give a feature map and a relevance map, respectively. Each element in the feature map is multiplied by the corresponding element in the feature map to give a normalized feature map that is passed (after using a rectified linear unit (RELU)) as input to the next layer of the net.

## 1.1. Data set of materials inside glassware vessel in a chemistry lab setting

In addition, a new data set of materials in transparent vessels in a chemistry laboratory setting is presented. A secondary goal of this data set is to supply training and testing sets for both the valve filter approach and other approaches using neural nets with an ROI input. The main purpose of the data set is to enable the training of a neural network in the task of recognizing the physical and chemical properties of materials inside glassware vessels in a laboratory setting. The handling of materials in glassware and other transparent vessels is the main activity in most chemistry laboratory work, and is essential for a wide range of methods used in materials research [5-9, 14]. The data set presented here includes a thousand images of chemicals in different physical phases in a chemistry laboratory setting. Each image in the data set is supplied with pixel-wise annotation (semantic segmentation) according to several sets of categories. The first level segments the image into the glassware region and the background region; the second layer segments the vessel region into filled and empty regions; the third layer segments the vessel contents into liquid and solid phases; and the final level gives exact categories for the phase of the material in the vessel, including liquid, solid, powder, foam, vapor, suspension and emulsion. The images from this data set were collected from YouTube channels dedicated to chemistry experiments (mainly RedNile, NudRage, ChemPlayer), and were manually labeled by Alexandra Emanuel and Mor Bismuth.

# 2. Related work

Following the success of Alexnet [4] in the Imagenet Classification Challenge in 2012, convolutional neural networks (CNN) have emerged as the leading methods for image classification, detection and segmentation [1, 2, 4, 15]. CNN work by convolving the image with a variety of learned filters to extract a feature map, which represents the distribution of a variety of features in the image [3]. These feature maps are then used as input for the next convolution layer. Each layer extracts higher-level features, which enable better abstraction of the image contents to be achieved. Since the convolution filters used in the CNN are learned, such nets can be trained to recognize various complex categories without human aid. However, training such nets requires a large number of training images annotated according to the specific task at hand. A number of data sets such as COCO Pascal and KADE20 are available, with thousands of pixel-level annotated images [10-12]. However, these sets contain general scenes and categories corresponding to everyday objects, and data sets for more specific cases are often not available. While CNNs were first demonstrated for image classification tasks [4] (hence assigning one category per image), they were soon adopted for the task of object detection [1, 16] and per-pixel-level category prediction (semantic segmentation) [2, 15].

## 2.1. Inserting attention region and ROI into neural nets

A standard approach for introducing a region of interest (ROI) as an input for neural nets is by assuming a rectangular ROI such as a bounding box around an object (Figure 4a). This rectangular ROI can then be cropped and used as a separate image that will be passed as input to the net; this type of approach was taken in RCNN for object detection [16] (Figure 4a). Later methods such Faster RCNN have accelerated this process by first passing the entire image through the net, and performing ROI cropping only on the feature map of the final layer [1]. One limitation of the cropping approach is that it assumes a rectangular ROI, whereas in cases such as the identification of the contents of glassware, the ROI takes the shape of the glassware (Figure 1.a). An alternative approach for using an ROI with an arbitrary shape is ignoring (zeroing out) any image feature extracted outside the ROI region [17] (Figure 4b).

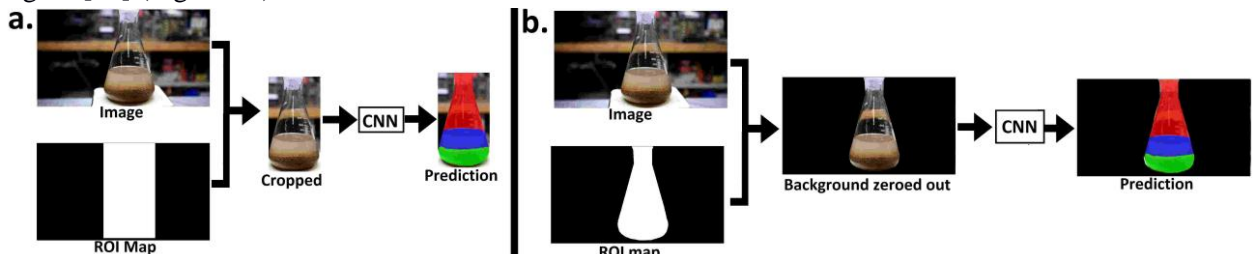

**Figure 4. Two methods for using convolutional neural nets (CNN) with ROI input: a) cropping the ROI region and using it as input for the net; b) blacking out features extracted outside the ROI region**

The problem with this approach is that it ignores background information that might be essential for the understanding of the context of many features within the ROI. For example, in the case shown in Figure 4b, without the background information it is impossible to determine whether the boundary line in the vessel corresponds to the material inside the vessel or to a background artifact. An alternative approach is to black out both the ROI and the background in different instances and to pass both instances as separate inputs to the net [18, 19]. Another approach that does not cause the loss of background information is called Direct Attention of Convolutional Neural Net (DCANN)[20]; this use the ROI map as another input for the net by concatenating it with inputs of specific layers of the net. The method suggested here also shares some similarities with attention map methods used with neural networks. These methods use semantic input such as image queries to generate an attention map for the image [21-23]. This map is then combined with the feature map for final prediction.

## 2.2. CNN for material classification and vessel contents recognition

Convolutional neural nets for material classification have also been suggested, and range from methods that classify the material of each pixel in the image using the full image information [24-26] to those which focus on a single image patch around the pixel [27-30]. In addition, several data sets were created for this task which contains labeled images or patches of images corresponding to different materials. These are either based on real world images (FMD, MINC) [24, 25] or on computer generated images (BTF) [27, 29]. Both the methods and the data sets for material recognition focus mainly on specific materials (metal/glass/wood/fabric) [24, 25] and not on the physical phase of the material (liquid/solid/powder/vapor/foam/emulsion), and cannot handle materials inside transparent vessels. The recognition of materials in transparent vessels has also been dealt with using classic computer vision methods. Most methods in this area focus on the recognition of a liquid level or phase boundary [5, 14, 30-35], in setting ranging from a bottle-filling module [32, 33] to liquid-liquid extraction [5, 6, 14]. These methods are usually very simple and focus on finding the phase boundary as a line or a curve. However, none of these methods deal with the recognition of the exact phases of the materials in the vessel. The settings of the images these methods can handle are also very simple, and assume that the vessel region is either constant in the image or that it was found using a method such as a template match or generalized Hough transform.

# 3. Valve filter approach to using nets with a ROI input

## 3.1. General approach

The approach suggested here is based on the idea that when analyzing an image divided into different known regions (such as the ROI and background) it is desirable to extract different information from different regions. Convolutional neural nets extract information from the image by convolving a set of filters with the image, whereby each filter extracts a different feature [3]. Extracting different information from different regions can, therefore, be done by applying different filters to different image regions. A more moderate approach is to apply the same set of filters to the entire image but to assume that different filters will have different relevance in different regions. Hence, some features might be more relevant to ROI regions, while others might be more relevant to background regions. For this, we will use a valve filter that inhibits specific features in a specific region of the image based on the ROI map (Figure 5). The valve filters act as a kind of valve that regulates the activation of the corresponding image filter in the different regions of the image. Applying the valve filters was done as shown in Figure 5, using the following steps: a) a set of filters are convolved with the image to generate a feature map; b) for each filter applied to the image, a corresponding valve filter exists that is convolved with the ROI map to give the relevance map; c) each element in the feature map is multiplied by the corresponding element in the relevance map to give a normalized feature map; d) the normalized featured map is used (after RELU) as input for the next layer. Hence, the relevance map is used to inhibit or enhance the activation of the features depending on whether they appear in the ROI or background region.

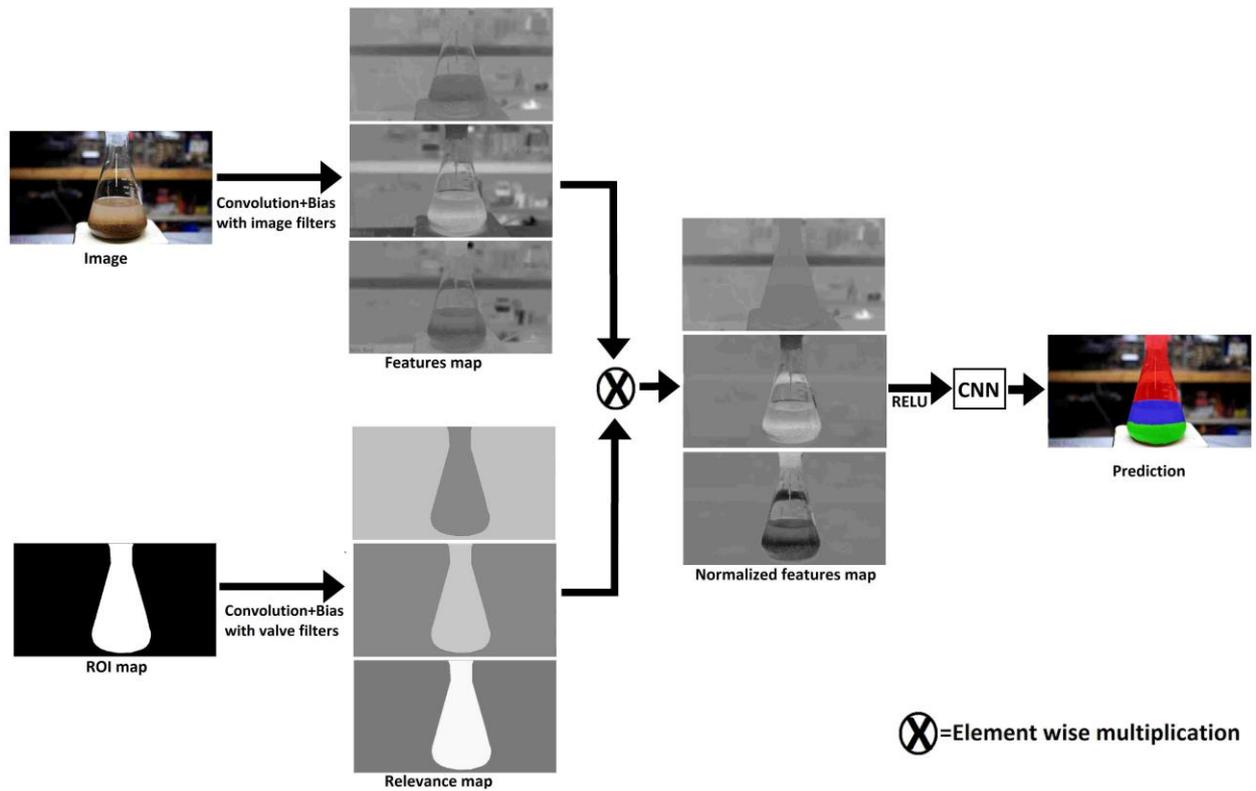

**Figure 5. Valve filter approach for the introduction of an ROI map as input to CNN. The image and ROI input are each passed through a separate convolution layer to give a feature map and a relevance map respectively. Each element in the feature map is multiplied by the corresponding element in the relevance map to give a normalized feature map that is passed (after a RELU) as input to the next layer of the net. Note that three maps are displayed for each step, although in practice 64 maps were extracted for each step**

### 3.2. Implementation details of the valve filters

The details of the implementation of the valve filters are given in Figure 5 and are described below:

1. The ROI map is inserted into the net along with the image. The ROI map is represented as a binary image, with pixels corresponding to ROI marked 1 and the remainder marked 0.
2. A set of image filters is convolved (with bias addition) with the image to give a feature map.
3. A set of valve filters convolved with the ROI map to give a relevance map with the same size and dimensions as the feature map (again with bias addition).
4. The feature map is multiplied elementwise by the relevance map. Hence, each element in the relevance map is multiplied by the corresponding element in the feature map to give a normalized feature map.
5. The normalized feature map is then passed through a RELU which zeroes out any negative map element. The output is used as input for the next layer of the net.

In this way, each valve filter acts as kind of a valve that regulates the activation of the corresponding image filter in different regions of the image. Hence, the valve filter will inhibit some filters in the background zone and others in the ROI zone (Figure 5). The weights of the valve filters are learned by the net in the same way as the image filters. The net therefore learns both the features and the region for which they are relevant. In the current implementation, the valve filter acts only on the first layer of the convolutional neural net, and the rest of the net remains unchanged. As a result, the valve can be used with most CNNs by modifying only the first layer.

## 3.3 Analyzing the valve filter relevance map

The relevance map generated by applying the valve filters on the ROI map is used to control the region of the image in which different image features will be active. Since the valve filters are learned by the net, it is interesting to examine the relevance map of several filters. Figure 6 displays the relevance map of several valve filters alongside the feature map of the corresponding image filter and the normalized feature map. Negative map regions are marked in red and positive regions in green; the intensity of the color corresponds to the absolute value. It can be seen from Figure 6 that the valve filters have indeed learned to direct specific filters to the ROI region (by giving low relevance to the background region), while other image filters are directed to the background region (by giving a low value to the ROI region). Interestingly, the relevance map is in some cases negative, meaning that it inverts the value of the corresponding feature map. In most cases, the relevance map is not binary, and simply scales out the responses of the various features in different regions. In addition, some relevance maps seem to have high values around the boundary of the ROI, suggesting that these regions are particularly important for some features. The feature map extracted from the image (Figure 6 center) suggests that these filters remain similar to those of standard CNNs, and focus on recognizing edges and textures. As expected, the normalized feature map (Figure 6 right), shows that for different filters either the ROI region or the background region was suppressed by the relevance map. This is clear proof that the net has learned to direct different filters to different image regions.

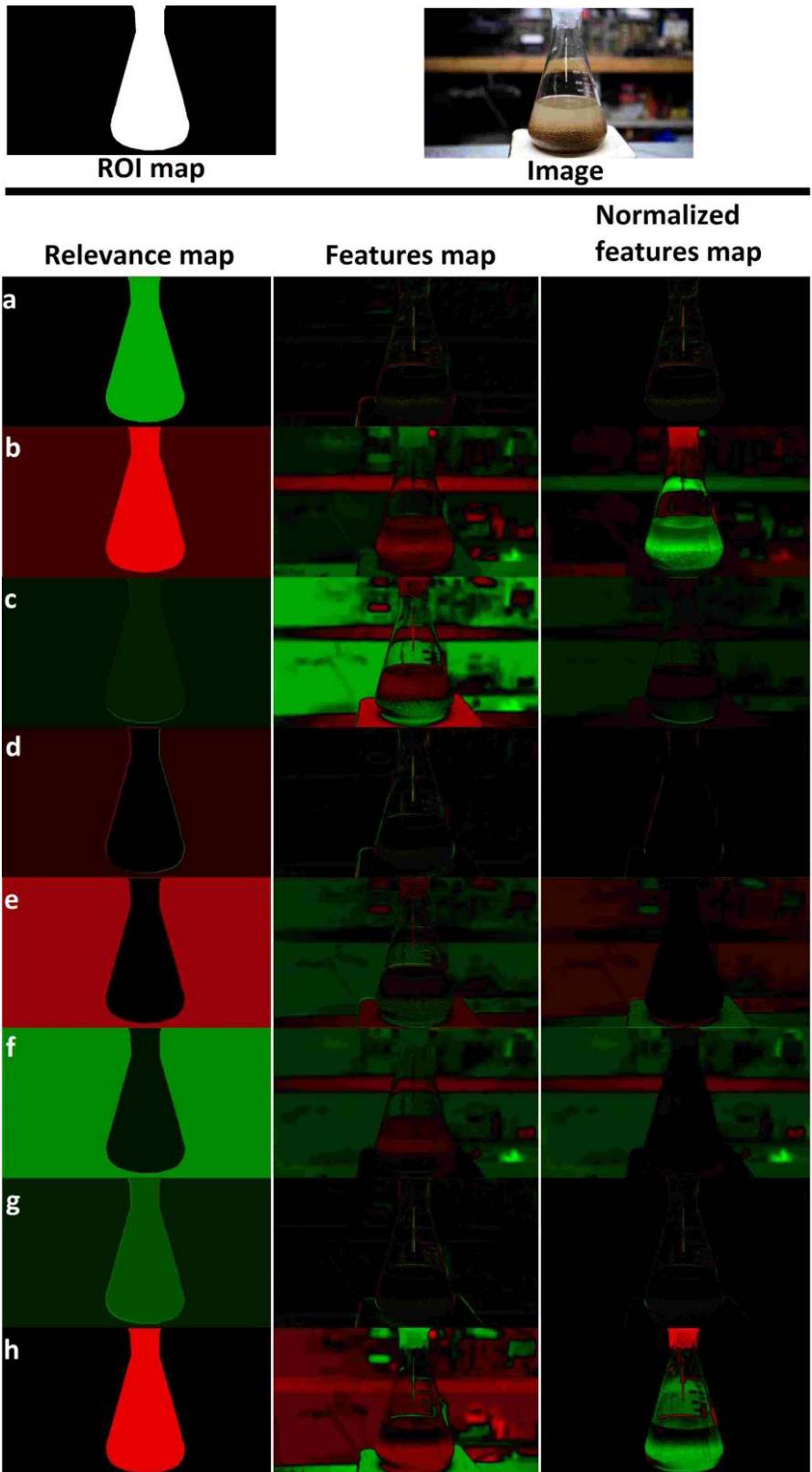

**Figure 6.** Relevance map (left) displayed alongside the corresponding feature map (center) and normalized feature map

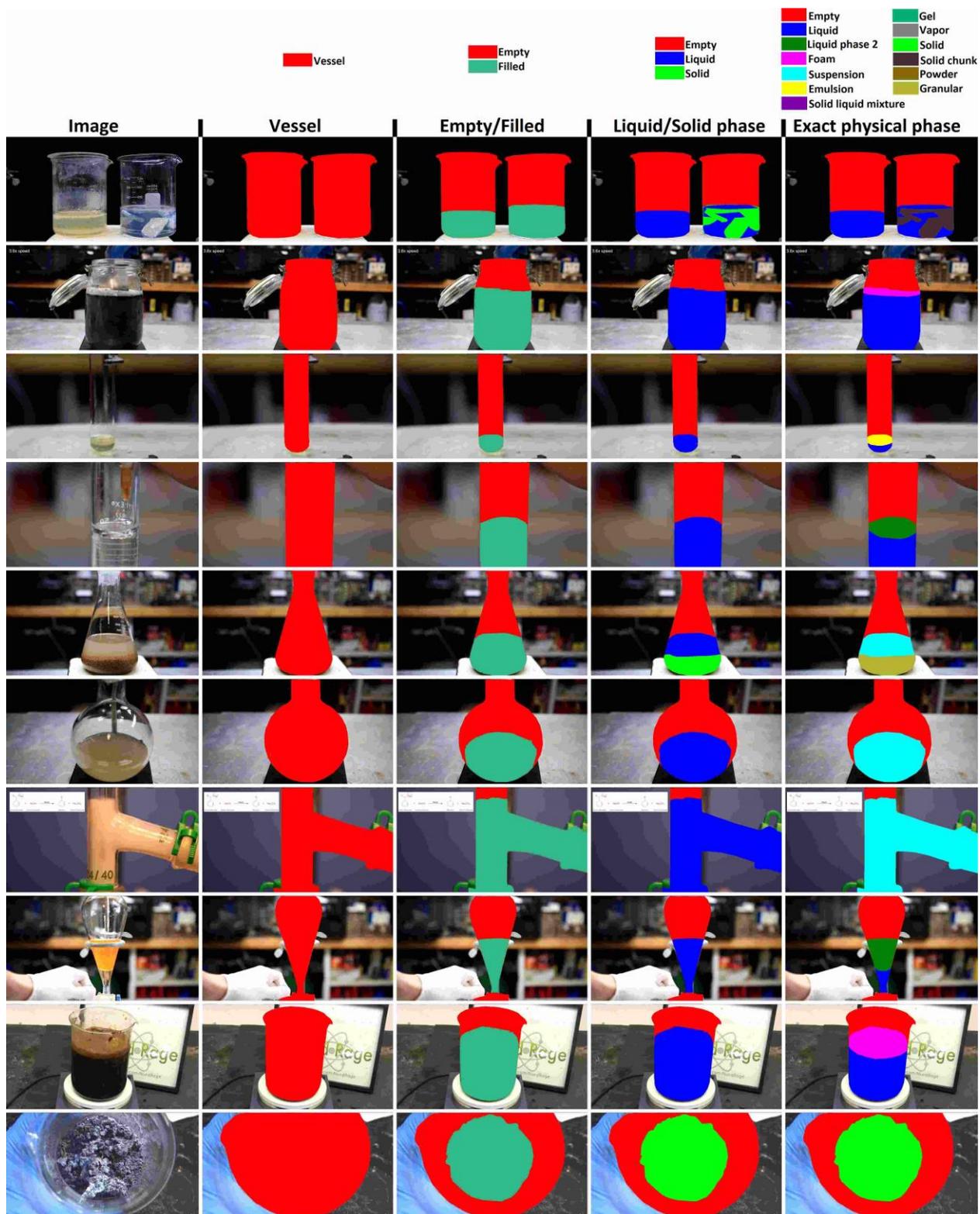

**Figure 7.** Images from the data set of materials in glass vessels and their annotation. Each column in the image displays a different level of annotation. For details see Section 4

# 4. Data set of materials in vessels

The handling of materials in glassware vessels is the main task in chemistry laboratory research [5-8, 14] as well as a large number of other activities. Visual recognition of the physical phase of the materials is essential for many methods ranging from a simple task such as fill-level evaluation to the identification of more complex properties such as solvation, precipitation, crystallization and phase separation. To help train neural nets for this task, a new data set was created. The data set contains a thousand images of materials, in different phases and involved in different chemical processes, in a laboratory setting. Each pixel in each image is labeled according to several layers of classification, as given below (Figure 7):

a. Vessel/Background: For each pixel assign value of one if it is part of the vessel and zero otherwise. This annotation was used as the ROI map for the valve filter method.

b. Filled/Empty: This is similar to the above, but also distinguishes between the filled and empty regions of the vessel. For each pixel, one of the following three values is assigned: 0 (background); 1 (empty vessel); or 2 (filled vessel).

c. Phase type: This is similar to the above but distinguishes between liquid and solid regions of the filled vessel. For each pixel, one of the following four values: 0 (background); 1 (empty vessel); 2 (liquid); or 3 (solid).

d. Fine-grained physical phase type: This is similar to the above but distinguishes between specific classes of physical phase. For each pixel, one of 15 values is assigned: 1 (background); 2 (empty vessel); 3 (liquid); 4 (liquid phase two, in the case where more than one phase of the liquid appears in the vessel); 5 (suspension); 6 (emulsion); 7 (foam); 8 (solid); 9 (gel); 10 (powder); 11 (granular); 12 (bulk); 13 (solid-liquid mixture); 14 (solid phase two, in the case where more than one phase of solid exists in the vessel): and 15 (vapor).

The annotations are given as images of the size of the original image, where the pixel value is the class number. The annotation of the vessel region (a) is used in the ROI input for the valve filter net .

## 4.1. Validation/testing set

The data set is divided into training and testing sets. The testing set is itself divided into two subsets; one contains images extracted from the same YouTube channels as the training set, and therefore was taken under similar conditions as the training images. The second subset contains images extracted from YouTube channels not included in the training set, and hence contains images taken under different conditions from those used to train the net.

## 4.2. Creating the data set

The creation of a large number of images with a variety of chemical processes and settings could have been a daunting task. Luckily, several YouTube channels dedicated to chemical experiments exist which offer high-quality footage of chemistry experiments. Thanks to these channels, including NurdRage, NileRed, ChemPlayer, it was possible to collect a large number of high-quality images in a short time. Pixel-wise annotation of these images was another challenging task, and was performed by Alexandra Emanuel and Mor Bismuth.

# 5. Methods and evaluation

## 5.1. Fully convolutional neural nets for semantic segmentation

Fully convolutional neural nets (FCN) are a type of neural net designed to predict, for each pixel in the image, the class of object or material to which it belongs. FCN was applied to the data set (Section 4) both with and without ROI input. For the implementation of FCN with ROI input, the ROI map was the region of the glassware vessels in the image taken from the data set annotation (Section 4a) or from the vessel region prediction of the net without the ROI (for hierarchical segmentation). The implementation of fully convolutional neural has been described in previous works, and will not be discussed here. The valve filters applied to the first layer of the net are described in Section 3. An implementation of this net using Tensorflow is supplied in the supporting material.

## 5.2 Method and evaluation

The valve filter approach was examined, along with several other approaches based on fully convolutional nets (FCN), for the task of semantic segmentation of the data set. The approaches tested are described below:
   a. Standard FCN without ROI input (Figure 2);
   b. FCN with valve filter and ROI input (Section 3);
   c. FCN with the ROI map as another channel to the input image;
   d. FCN with zeroing out of features outside the ROI (Figure 4b).

The results were evaluated using the test set (Section 4.1), and are given in Tables 1 and 2. The method of evaluation was pixel-wise intersection over union (IOU) of the labels predicted by the net and ground truth labels of the test set (Section 4). Example segmentations made by the valve filter method are shown in Figure 9. Example segmentations made by the standard FCN net without ROI input are shown in Figure 10.

## 5.3. Multi-class vs. single-class prediction

Training neural nets to annotate images using the data set can be done by teaching the net to predict a single set of categories and using a different net for each set of categories (Figure 8a). An alternative approach is to teach a single net to make predictions for several layers of categories simultaneously (Figure 8b). Training a different net for predicting each set of categories (Figure 8a) give superior results when compared to training one net with multilevel prediction (Figure 8b). However, this approach is much more time consuming in both the training and inference stages.

## 5.4. One step segmentation vs. hierarchical segmentation

The ROI input used by the valve filters net in this cases is the glass vessel region in the image. One option is to assume that this region is pre given and use the ground truth vessel region from the data set as the ROI. However, if the ROI region in the image is not known there are two alternatives: One option is to use a standard net for finding both the vessel (ROI) region and its contents in a single step (Figure 2). The second option is to use a hierarchical approach in which one net to find the vessel region in the image and use this region as the ROI input for a second net that will identify the content of the ROI (in this case the materials inside the vessel). While both approaches receive only the image as an input, the hierarchical approach gave far better results (Tables 1 and 2).

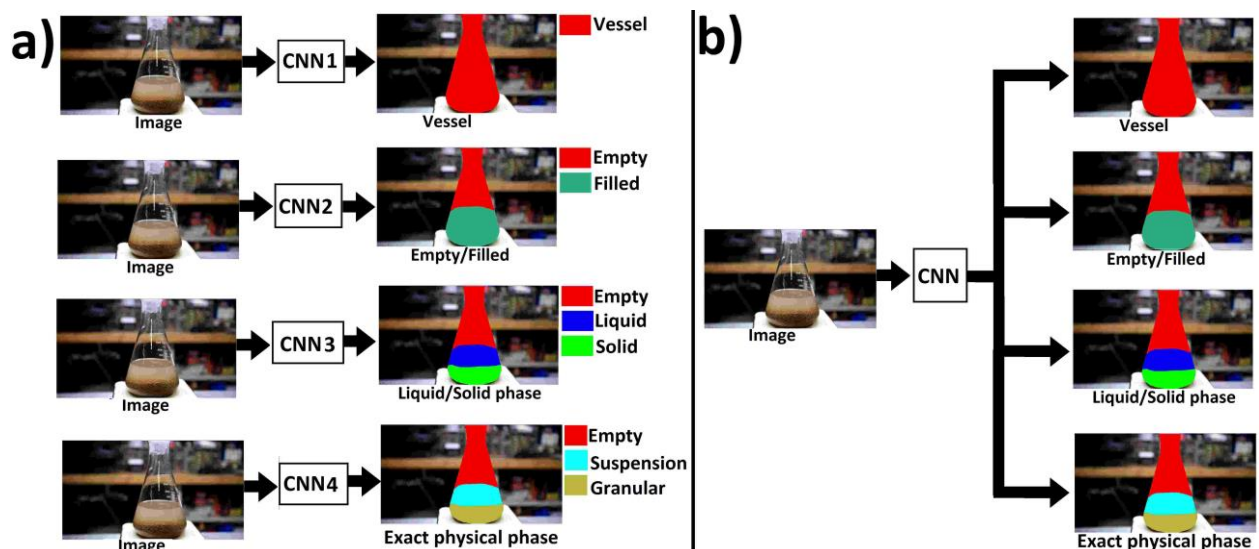

**Figure 8.** Two approaches for multi-level pixel-wise classification: a) apply a different net for each level of classification; b) apply a single net with multi-level classification.

# 6. Results and discussion

## 6.1. General discussion

It can be seen from Tables 1 and 2 and Figures 9 and 10 that the valve filter approach outperformed all the other methods by a large margin. Even when ignoring predictions outside of the vessel region, the valve filter method outperforms all the other methods examined. This suggests that the use of the valve approach in processing the ROI map allows the net to focus learning on recognition of the vessel's contents. The network without ROI input was forced to focus more resources on learning, in order to detect the glassware vessel region. All nets performed poorly in the exact categorization of material phases (Tables 1 and 2). This can be attributed to two factors: a) a lack of training data; and b) the more specific the material phase label, the less likely it is to appear in the data set. Since the majority of the materials appearing in the data set are liquids, the other phases of matter appear rather seldom, and there are therefore an insufficient number of examples in the data set to allow for proper training. In addition, distinguishing between different phases of matter from a single image is a hard, if not impossible, task in some cases (for example, distinguishing between a emulsion and a suspension); even the human labelers were unable to decide in some cases.

## 6.2. Analysis of standard FCN semantic segmentation with no ROI input

In prediction without the use of the vessel region as ROI input, the task is particularly hard, both because of a relatively small training data set and the fact that glassware and other transparent vessels are rather hard to recognize due to their transparency and strong reflectance. However, the net still achieves reasonable results and predictions (Table 1 and Figure 10) when tested on images taken under the same conditions as the training set (images taken from the same YouTube channel as the training data). However, when tested on images taken in different settings from the training set, this net performed poorly (Table 2). This suggests that while it is possible to achieve good results without inserting the vessel region as ROI input into the net, this type of net requires either far more training data or training data that were taken under conditions similar to those in the use case.

## 6.3. Results of net with ROI mask as input

The results in Tables 1 and 2 and Figure 9 show that using the vessel region as ROI input to the net gives far better results compared to semantic segmentation without ROI input. This is true even when the vessel region used as the ROI, was found by another net (Hierarchical segmentation, Figure 1.b). The valve filter approach outperformed all the other approaches by a large margin, even when ignoring false prediction outside the vessel region. This implies that the valve filter allows the net to better focus the learning process inside the ROI region. Comparing the valve filter approach to the approach based on zeroing-out of the background region (Figure 4b) shows that both systems learned to use the ROI to find the vessel region. However, the zeroing-out approach achieved far worse results in identifying the vessel contents (Tables 1 and 2). This supports the idea that completely ignoring the background region of the image discards important information and erodes the accuracy of the net. Interestingly, the method of introducing the ROI/background map as another channel to the net input gave imperfect results, even in the task of predicting the vessel region (Tables 1 and 2). This is surprising, since the vessel region is simply the ROI input. This implies that, in this approach, the net failed to transfer the information about the vessel's location from the input to the higher layers of the net. Entering the ROI map as input to higher levels of the net, as in the DCANN approach, might solve this problem; however, this will require more complex modifications of the nets that will have to be adapted for each type of net.

Table 1: Intersection over union testing (easy) same testing and training setting

| | Single label prediction[1] | | | Multi label prediction[2] | | | |
|---|---|---|---|---|---|---|---|
| | Valve filters[3] | No ROI input[4] | Hierarchical Segmentation valve filters[5] | Valve filters[3] | No ROI input[4] | Background blacked out[6] | ROI Concatenated to input[7] |
| **Vessel region** | | | | | | | |
| Background | 100% | 94% | 94% | 100% | 86% | 99% | 98% |
| Vessel | 99% | 84% | 83% | 99% | 67% | 98% | 95% |
| **Fill level** | | | | | | | |
| Background | 100% | 93% | 94% | 100% | 86% | 99% | 98% |
| Empty | 82% | 65% | 67% | 82% | 49% | 79% | 74% |
| Filled | 82% | 70% | 73% | 82% | 52% | 79% | 77% |
| **Solid/Liquid** | | | | | | | |
| Background | 100% | 93% | 94% | 100% | 86% | 99% | 98% |
| Empty | 84% | 62% | 68% | 82% | 48% | 79% | 74% |
| Liquid | 78% | 59% | 70% | 74% | 47% | 65% | 67% |
| Solid | 62% | 47% | 60% | 42% | 36% | 28% | 36% |
| **Exact physical phase** | | | | | | | |
| Background | 100% | 93% | 94% | 100% | 87% | 99% | 98% |
| Vessel | 80% | 60% | 66% | 81% | 47% | 78% | 73% |
| Liquid | 58% | 44% | 51% | 54% | 39% | 51% | 51% |
| Liquid Phase two | 22% | 17% | 22% | 17% | 2% | 13% | 12% |
| Suspension | 29% | 27% | 29% | 28% | 7% | 21% | 29% |
| Emulsion | 1% | 2% | 1% | 1% | 0% | 5% | 2% |
| Foam | 5% | 6% | 7% | 7% | 1% | 1% | 3% |
| Solid | 9% | 0% | 9% | 6% | 13% | 7% | 10% |
| Gel | 0% | 0% | 0% | 0% | 0% | 0% | 0% |
| Powder | 27% | 25% | 28% | 15% | 11% | 12% | 26% |
| Granular | 29% | 28% | 25% | 46% | 15% | 15% | 33% |
| Bulk | 8% | 2% | 9% | 4% | 0% | 1% | 1% |
| Solid liquid mixture | 1% | 1% | 2% | 1% | 0% | 0% | 1% |

[1] Use a net to predict a single label per pixel. Use a different net for every set of labels (Section 5.3, Figure 8.a)

[2] Use a net to predict multiple labels per pixel. Use a single net generate multiple sets of labels per pixel (Figure 8.b)

[3] Valve filter network with ground truth vessel region as ROI input (Figure 5)

[4] Use standard FCN with only the image and no ROI map as input (One Step prediction, Section 5.4, Figure 2)

[5] Use valve filters net where the ROI input is the vessel region predicted by standard FCN (Section 5.4, Figure 1.b)

[6] Black out features outside the ROI region (Figure 4.b)

[7] Concatenate ROI map to image (Section 2.1)

### Table 2: Intersection over union testing (hard) different testing and training settings

| | Single label prediction[1] | | | Multi label prediction[2] | | | |
|---|---|---|---|---|---|---|---|
| | Valve filters[3] | No ROI input[4] | Hierarchical Segmentation valve filters[5] | Valve filters[3] | No ROI input[4] | Background blacked out[6] | ROI Concatenated to input[7] |
| **Vessel region** | | | | | | | |
| Background | 100% | 91% | 91% | 100% | 80% | 99% | 96% |
| Vessel | 99% | 74% | 74% | 99% | 58% | 98% | 90% |
| **Fill level** | | | | | | | |
| Background | 100% | 89% | 91% | 100% | 81% | 99% | 96% |
| Empty | 73% | 47% | 50% | 71% | 37% | 67% | 62% |
| Filled | 80% | 56% | 63% | 77% | 45% | 75% | 71% |
| **Solid/Liquid** | | | | | | | |
| Background | 100% | 88% | 91% | 99% | 81% | 99% | 96% |
| Empty | 74% | 42% | 51% | 71% | 37% | 67% | 62% |
| Liquid | 74% | 54% | 58% | 72% | 40% | 62% | 63% |
| Solid | 37% | 21% | 31% | 19% | 20% | 11% | 12% |
| **Exact physical phase** | | | | | | | |
| Background | 100% | 87% | 91% | 99% | 81% | 99% | 96% |
| Vessel | 68% | 41% | 48% | 70% | 35% | 65% | 61% |
| Liquid | 55% | 38% | 43% | 53% | 35% | 48% | 44% |
| Liquid Phase two | 2% | 1% | 2% | 3% | 0% | 1% | 0% |
| Suspension | 22% | 12% | 18% | 16% | 6% | 13% | 18% |
| Emulsion | 18% | 0% | 3% | 6% | 0% | 6% | 0% |
| Foam | 5% | 2% | 6% | 9% | 1% | 4% | 3% |
| Solid | 31% | 0% | 29% | 12% | 13% | 3% | 6% |
| Gel | 4% | 0% | 0% | 0% | 0% | 0% | 0% |
| Powder | 0% | 6% | 6% | 2% | 4% | 6% | 8% |
| Granular | 0% | 0% | 0% | 0% | 8% | 0% | 3% |
| Bulk | 0% | 0% | 0% | 0% | 0% | 0% | 0% |
| Solid liquid mixture | 0% | 0% | 0% | 0% | 0% | 0% | 0% |

[1] Use a net to predict a single label per pixel. Use a different net for every set of labels (Section 5.3, Figure 8.a)

[2] Use a net to predict multiple labels per pixel. Use a single net generate multiple sets of labels per pixel (Figure 8.b)

[3] Valve filter network with ground truth vessel region as ROI input (Figure 5)

[4] Use standard FCN with only the image and no ROI map as input (One Step prediction, Section 5.4, Figure 2)

[5] Use valve filters net where the ROI input is the vessel region predicted by standard FCN (Section 5.4, Figure 1.b)

[6] Black out features outside the ROI region (Figure 4.b)

[7] Concatenate ROI map to image (Section 2.1)

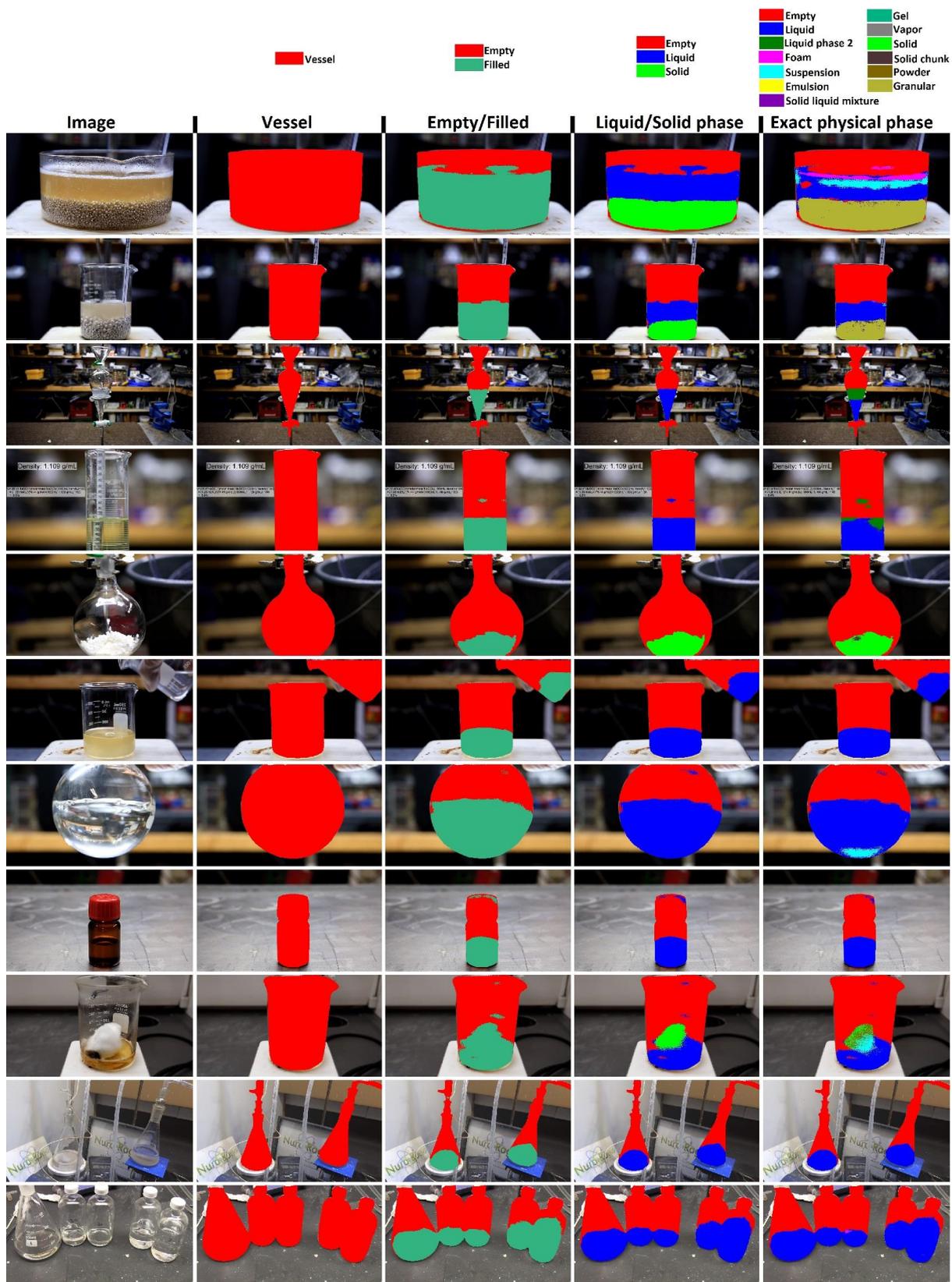

Figure 9. Some results of the valve filter method for the test set of materials in glass vessels (Section 4.1). The vessel region was used as the ROI input (Section 4). Each column gives predictions for a different classification level. The leftmost image was the net input

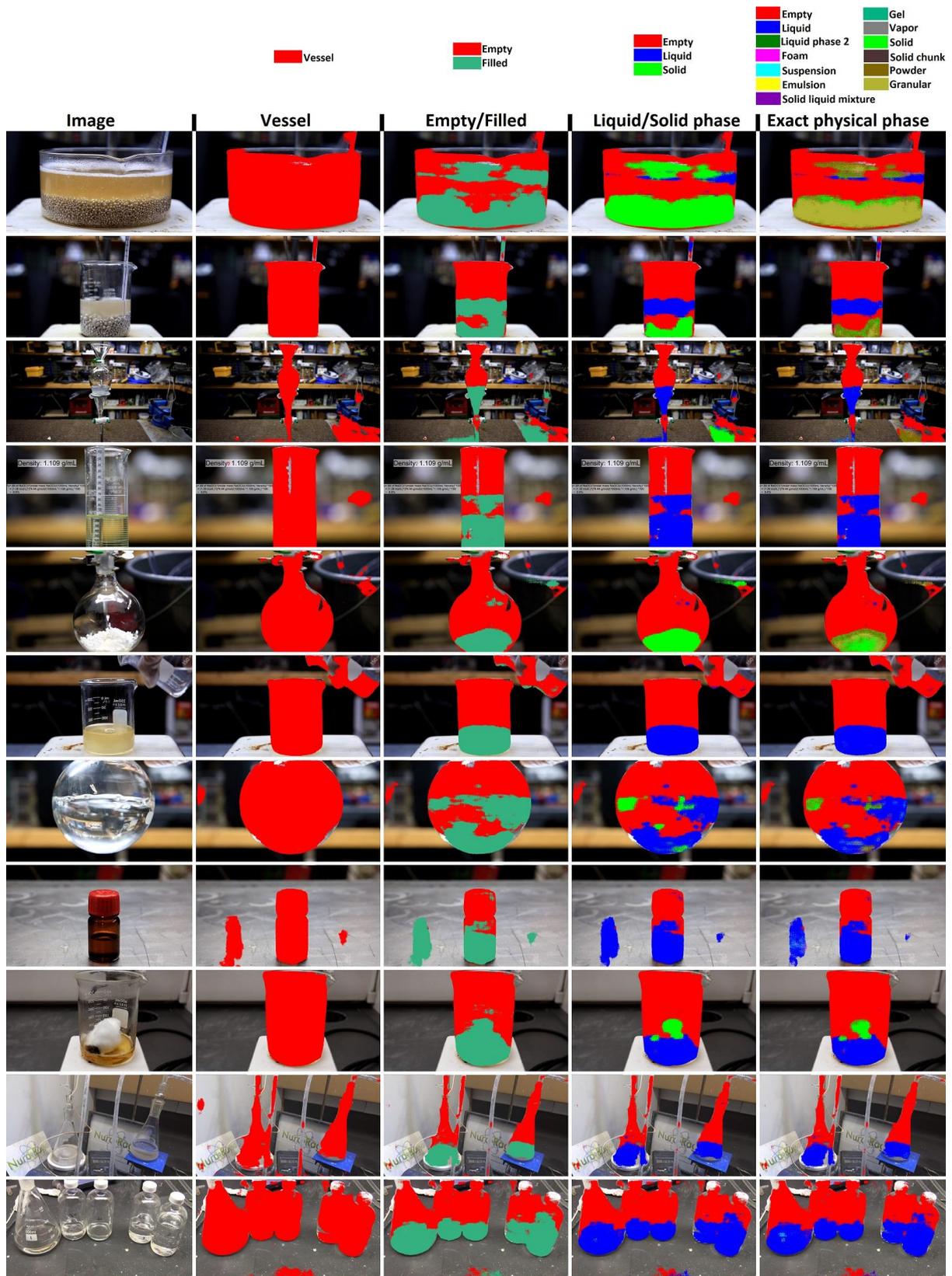

Figure 10. Some results of a fully convolutional neural net with no ROI input for the materials in vessels test set (Section 4.1). Each column gives predictions for a different classification level. The leftmost image was the net input

# 7. Conclusion

The two goals of this work were: i) to explore a new method of inserting arbitrarily shaped region of interest (ROI) inputs to neural networks; and ii) to exploring this approach in the recognition of materials inside transparent vessels in a chemistry laboratory setting. The insertion of an ROI as input for the net can be useful when the image is already segmented into foreground and background areas. The test case used here was to identify the contents of a glassware vessel in the image, where the vessel region was already known. The ROI map was introduced as an input to the net using a valve filter approach. This approach is based on making different convolutional filters focus on different regions of the image, based on the ROI map. This method significantly outperformed nets without an ROI input mask and nets in which the ROI input was added using other methods. The second goal of this work was to examine the task of recognizing material phases inside a glassware vessel in a chemistry laboratory setting. This task is essential in laboratory-based chemistry research, as well as in other fields in which materials are handled in glassware vessels. For this task, a new data set was presented of a thousand annotated images of materials and chemical process in a laboratory setting. The nets trained on this data set gave good results in predicting empty and filled regions and the liquid phases of matter. However, it gave poor results in the recognition of the exact physical phases of materials, suggesting that further work and a larger data set are needed.

# 8. Thanks

Thanks are due to Alexandra Emanuel and Mor Bismuth for their work on the labeling of the data set, and to the creators of the Youtube channels NileRed, NurdeRage and ChemPlayer for allowing the use of frames from their videos in the creation of the data set and this paper.

# 9. Supporting material

The dataset for materials in vessels (Section 4) can be found in: https://github.com/sagieppel/Materials-in-Vessels-data-set

Code for the valve filter approach (Section 3) can be found in: https://github.com/sagieppel/Focusing-attention-of-Fully-convolutional-neural-networks-on-Region-of-interest-ROI-input-map-

Code for FCN with no ROI used for the net can be downloaded from: https://github.com/sagieppel/Fully-convolutional-neural-network-FCN-for-semantic-segmentation-Tensorflow-implementation